\documentclass{article}

\usepackage[preprint]{neurips_2024}

\usepackage[utf8]{inputenc} % allow utf-8 input
\usepackage[T1]{fontenc}    % use 8-bit T1 fonts
\usepackage{hyperref}       % hyperlinks
\usepackage{url}            % simple URL typesetting
\usepackage{booktabs}       % professional-quality tables
\usepackage{amsfonts}       % blackboard math symbols
\usepackage{nicefrac}       % compact symbols for 1/2, etc.
\usepackage{microtype}      % microtypography
\usepackage{xcolor}         % colors
\usepackage{graphicx}
\usepackage{amsmath}
\usepackage{subcaption}

\title{Modularity in Transformers: Investigating Neuron Separability \& Specialization}

\author{%
  Nicholas Pochinkov \\ %\thanks{Use footnote for providing further information about author (webpage, alternative address)---\emph{not} for acknowledging funding agencies.} \\
  Independent \\
  Dublin, Ireland \\
  \texttt{work@nicky.pro} \\
  \And
  Thomas Jones \\
  Independent \\
  London, UK \\
  \And
  Mohammed Rashidur Rahman\\
  Independent \\
  London, UK \\
}

\begin{document}

\maketitle

\begin{abstract}
Transformer models are increasingly prevalent in various applications, yet our understanding of their internal workings remains limited. This paper investigates the modularity and task specialization of neurons within transformer architectures, focusing on both vision (ViT) and language (Mistral 7B) models. Using a combination of selective pruning and MoEfication clustering techniques, we analyze the overlap and specialization of neurons across different tasks and data subsets. Our findings reveal evidence of task-specific neuron clusters, with varying degrees of overlap between related tasks. We observe that neuron importance patterns persist to some extent even in randomly initialized models, suggesting an inherent structure that training refines. Additionally, we find that neuron clusters identified through MoEfication correspond more strongly to task-specific neurons in earlier and later layers of the models. This work contributes to a more nuanced understanding of transformer internals and offers insights into potential avenues for improving model interpretability and efficiency.
\end{abstract}

\section{Introduction}

As transformer neural networks continue to scale up and demonstrate improvements across various applications, their integration into more areas of technology becomes increasingly widespread. Despite these advances, our fundamental understanding of how these models operate remains limited \citep{unsolved-ml-safety}. This gap in understanding poses challenges for the reliability and trustworthiness of these models, particularly in high-stakes situations.

To address this, our study draws inspiration from bottom-up circuits-based \citep{circuits-induction-heads} and top-down linear direction-based analyses \citep{representation-engineering}, as well as from modularity research \citep{modular-deep-learning,emergent-modularity,modularity-clusterability}. We aim to explore the specialization of different components within neural networks in performing distinct tasks and querying different knowledge bases, defining modularity in this context as having two key traits: 1) A high degree of separability between components tasked with different functions, and 2) Positional locality of these components. Our focus primarily lies on the identification of distinct components within these networks and understanding their roles in processing different types of information.

\section{Method}

\subsection{Pre-Trained Models}

We work with a Vision Transformer, ViT-Base-224 \citep{vision-transformer-google-2021}.
as well as with a causal language model,
Mistral 7B instruct v0.2 \citep{mistral}.

The models are accessed via the HuggingFace transformer library \citep{DBLP:journals/corr/abs-2211-09085}. For each model, we perform the analysis on both the pre-trained model (called the 'trained model') and a single seed of a randomly initialized model (called the 'random model').

The neurons of study are found post-activation in the MLP layers, that is, between the $W_{in}$ and $W_{out}$ weight matrices, after the activation function (see \citep{geva-promoting-concepts}).

\subsection{Datasets and Tasks}
For ViT, the task of study is to perform image classification on different classes of Cifar100 \cite{cifar} fine-tuned model as a `retained' baseline. For the `forget' datasets, each super-class of Cifar100 (Cifar20) is used as a target. 

For text models, we investigate next-token prediction on various categories of text as possible niches, as well as subject-split MMLU question answering \cite{mmlu}. For a broad baseline of model activations, ElutherAI's `The Pile' \cite{pile}\footnote{
We use a subset of The Pile on HuggingFace with most copyrighted material removed.
} is used.
For the specialised datasets, use the individual subsets of the pile, as well as some some additional data sets.
We use some scientific-subject specific instruction datasets on specific topics from \cite{camel-ai-datasets-physics-bio-chem-etc}, including `Biology', `Physics', `Chemistry'.
We also use `CodeParrot GitHub Code' \cite{tunstall2022natural}, or `Code' for short.

\subsection{Neuron Selection and Neuron Clustering}\label{sec:neuron-scoring}

\subsubsection{Neuron selection}

We select neurons using a method analogous to the Selective Pruning approach \citep{pochinkov2024selectivepruning}. Given two datasets: a 'reference' dataset $D_r$ and 'unlearned' dataset $D_u$, we score the neurons $S_{n^l_i}$ by getting the ratio of mean absolute deviations between the two datasets, as seen in Equation \ref{eq:mean_abs}:
\begin{equation}\label{eq:mean_abs}
S_{n^l_i} := \frac{
\frac{1}{|D_r|} \sum_{x_i\in  D_r} | n^l_i ( x_i ) |
}{
\epsilon + \frac{1}{|D_u|} \sum_{x_j \in D_u}   | n^l_i ( x_j ) | 
}
\end{equation}

where $n^l_i(x_j)$ is the activation of neuron $i$ in layer $l$ for input $x_j$.

We then "select" a subset of neurons as being relevant to some data subset $D_q$ by taking a percentage of neurons with the highest scores. The percentage of neurons selected for each data subset is determined by aiming for a specific drop in top1 accuracy when these neurons are removed.

For ViT, the 'retained' baseline is Cifar100 \citep{cifar}, and the 'unlearned' data subsets are those of Cifar20. For Mistral, the retained baseline is ElutherAI's 'The Pile' \citep{pile}, with the subsets being the different Pile subsets and the additional specialized datasets mentioned earlier.

\subsubsection{MoEfication clustering}

We employ 'Mixture of Experts'-ification \citep{moefication} as a method for grouping neurons into clusters. This approach involves taking the input weights $W_{in}$ for an MLP layer and treating its columns as a collection of vectors. We then perform balanced k-means clustering \citep{balancedkmeans} on these vectors:
\begin{equation}
\{C_1, ..., C_k\} = \text{BalancedKMeans}(W_{in}^T, k)
\end{equation}

where $C_i$ represents the $i$-th cluster of neurons. This effectively splits each MLP layer into $k$ balanced 'mixture of experts' clusters. The intuition is to group neurons that are often activated simultaneously, allowing for efficient expert selection during inference.

For ViT, we use $k=96$ as in the original paper. For Mistral, $k=128$ is used for even divisibility of layers. We perform this clustering on both pre-trained and randomly-initialized models, providing a basis for comparing the emergent structure in trained networks against random initializations.

\section{Class-wise Unlearning Performance Evaluation}

\subsection{ViT Performance}

We assess the performance of ViT across all Cifar20 classes to determine if and how the performance on other classes is affected by the deactivation of neurons associated with a particular class. We uncover that between classes with commonalities, there often seem to exist shared neural pathways.

\begin{figure}[h]
    \includegraphics[width=0.55\linewidth]{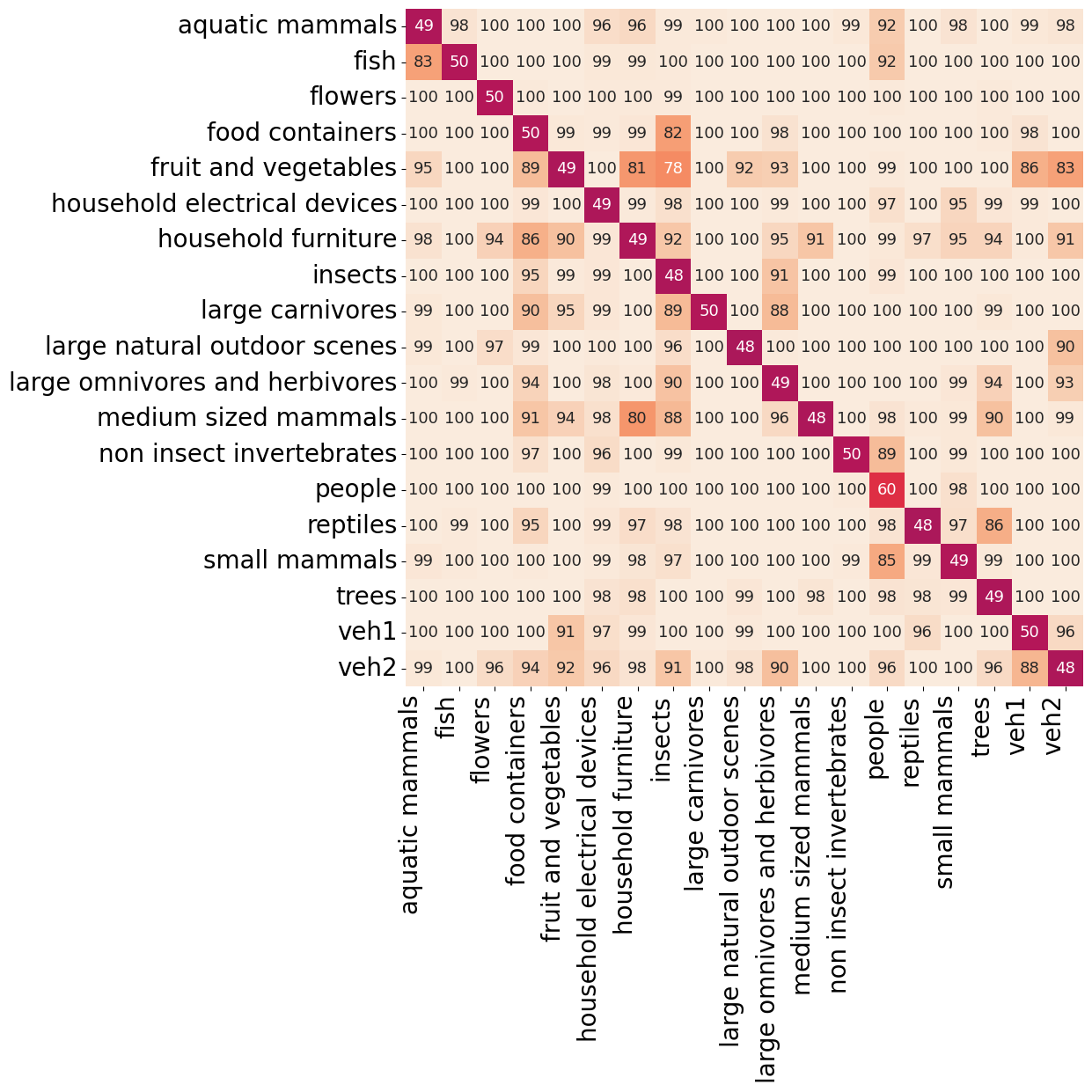}
    \includegraphics[width=0.425\linewidth]{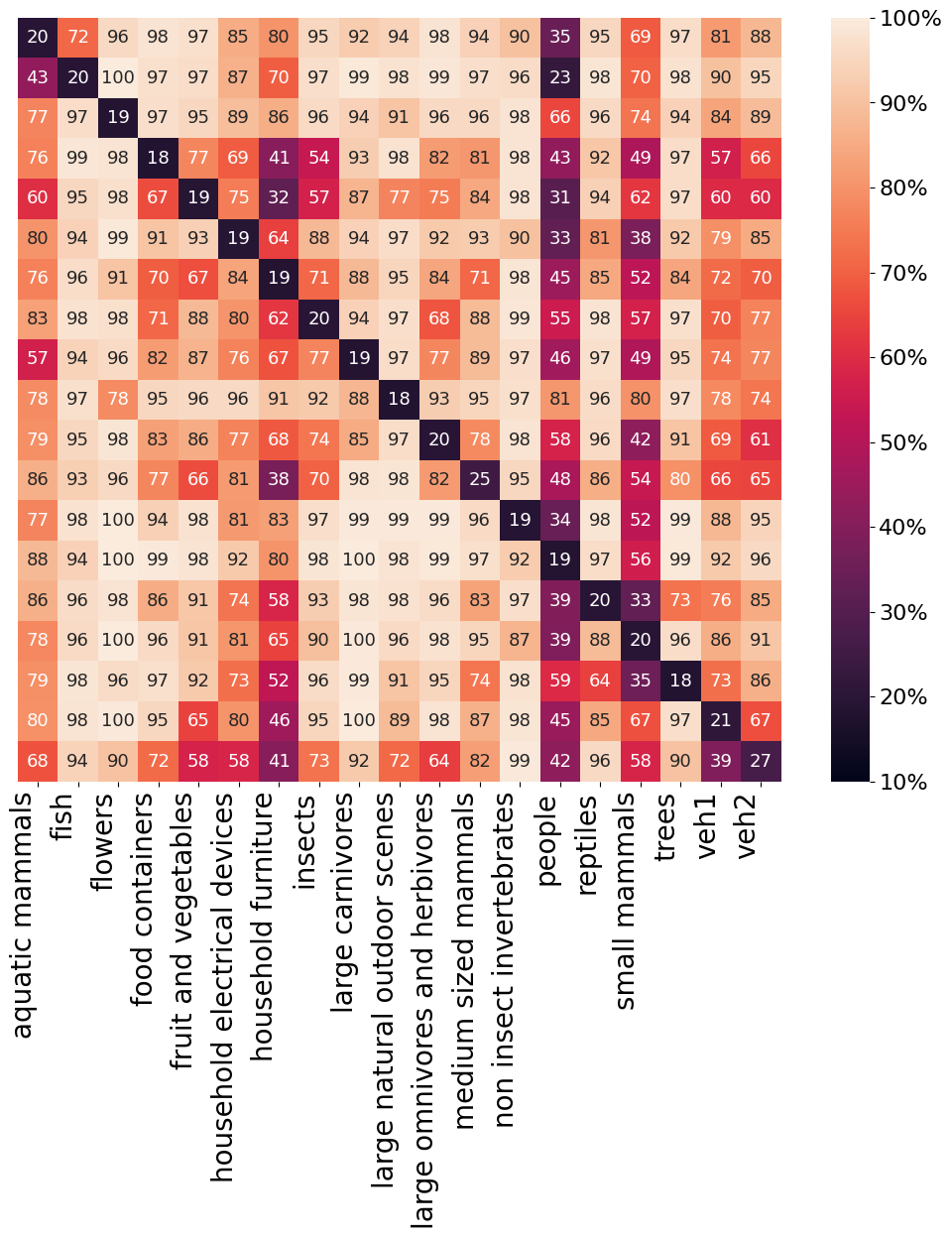}
    \caption{Relative Top1 \% image classification accuracy compared to baseline for ViT in Cifar100, after selectively pruning MLP neurons to get a target drop in performance of 50\% (left) and 80\% (right) for the 'unlearned' class (shown as the vertical strips).}
    \label{fig:cifar-eval-intetsections}
\end{figure}

We notice that the MLP neurons selectively pruned in ViT for classes such as 'insects' and 'non-insect invertebrates' seem to have relatively high overlap, and classes such as 'fish', 'aquatic mammals', and 'reptiles' also have correlated decreases in performance. We also observe that for two classes $A$ and $B$, the effect of unlearning class $A$ on the performance of class $B$ is not symmetric to unlearning class $B$ affecting the performance of class $A$, though these effects are correlated.

As we approach attempting to get an 80\% drop in top1 accuracy, the selectivity seems to significantly decrease, likely because the attention layers, which we do not modify, provide too strong a signal.

\subsection{Mistral Performance}

For Mistral, we investigate the degree to which a drop in performance in each class causes a correlated drop in performance across all other dataset tasks. We choose an arbitrary goal of unlearning by selective pruning until achieving a 20\% drop in Top1 next-token prediction accuracy in each respective unlearned text subset.

\begin{figure}
    \centering
    \includegraphics[width=0.95\linewidth]{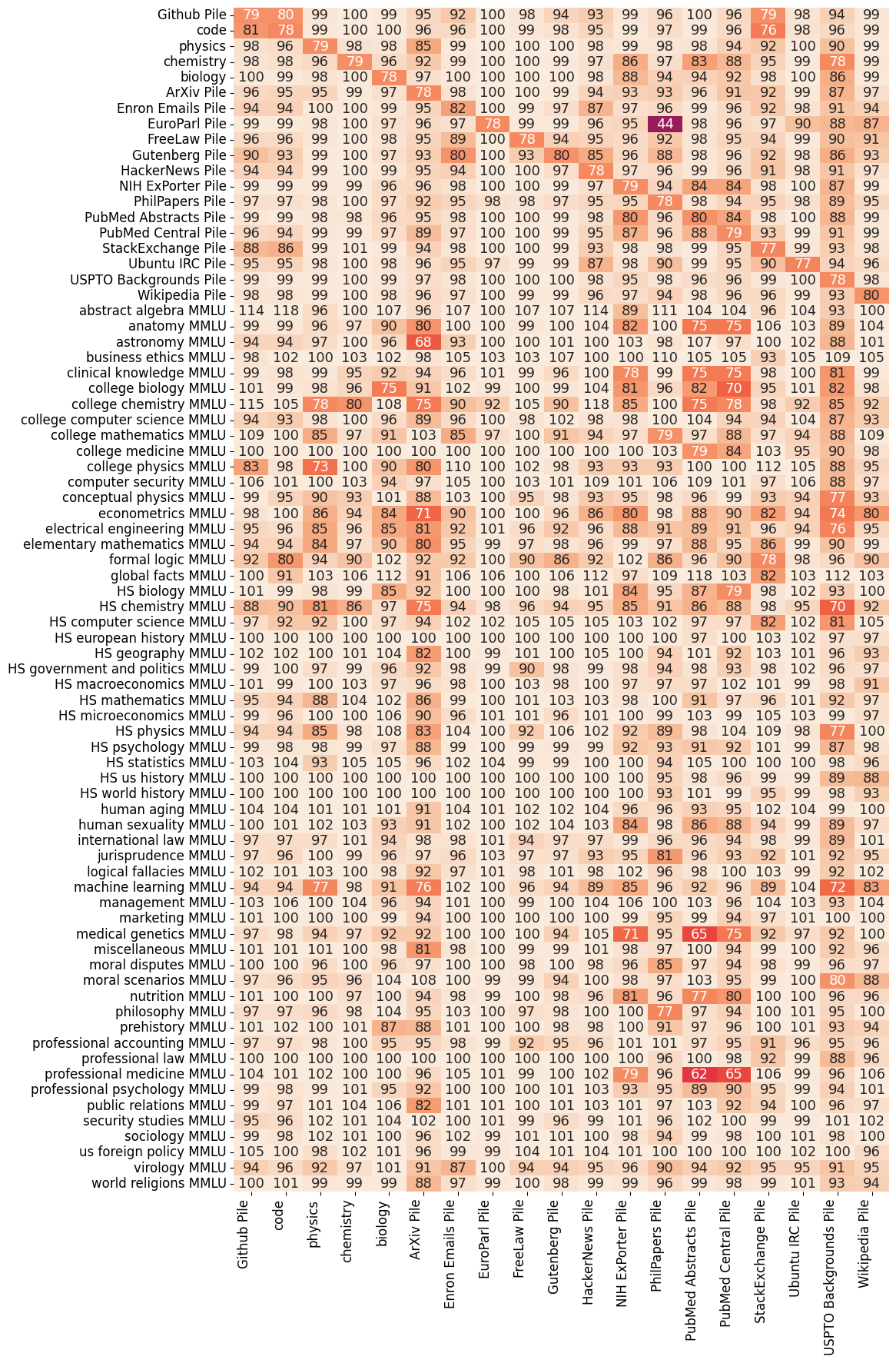}
    \caption{Drop in Top1 accuracy in Mistral 7B for next-token-prediction datasets as well as zero-shot subject-split MMLU question answering. We target a relative drop in performance of 20\% for the class being 'unlearned', and each vertical strip shows the effect on the other data subsets.}
    \label{fig:mmlu-chungus}
\end{figure}

In Figure \ref{fig:mmlu-chungus}, we observe that in most subsets of data that one would intuitively associate, such as unlearning 'biology' leading to a drop in MMLU 'High School biology'. Notably, we see that unlearning 'PubMed central' leads to significant drops in MMLU 'professional medicine', 'Medical Genetics', 'Clinical Knowledge', 'Anatomy', 'College Biology', and 'College Chemistry'. Additionally, we observe that unlearning 'ArXiv' leads to corresponding drops in MMLU subjects.

We also note some 'failure' cases:
\begin{itemize}
    \item 'EuroParl' and 'FreeLaw' seem to have relatively modest effects on the accuracy of other subsets, including seemingly related MMLU topics such as 'HighSchool Government \& Politics' and 'Professional Law'.
    \item In some cases, performance seems to increase, including up to 118\% of baseline in MMLU 'Abstract Algebra' after performing the unlearning of Code.
    \item Unlearning 'PhilPapers' seems to have an outsized negative impact on 'EuroParl' performance, significantly more so than even the 'EuroParl' unlearning has.
\end{itemize}

These results highlight that the Selective Pruning method seems to perform reasonably well at identifying relatively general class-scale features of models, despite being relatively simplistic in looking at the activations. However, there is potential room for improvement with better methods and datasets. It does, nonetheless, provide a promising glimpse of some kind of interpretable modularity.

\section{Intersection Analysis}

\subsection{Intersection between Class Selected Neurons}

We analyze the degree to which selected neurons overlap between different classes or data subsets. For some set of selected neurons $N_i$ and $N_j$ for classes $i$ and $j$ respectively, we calculate $|N_i \cap N_j| / |N_i|$.

Each class of neurons amounts to a total of $12.5 \pm 6.0\%$ of neurons in ViT, and $3.4 \pm 2.6\%$ of neurons in Mistral.

\begin{figure}[h]
    \includegraphics[width=0.588\linewidth]{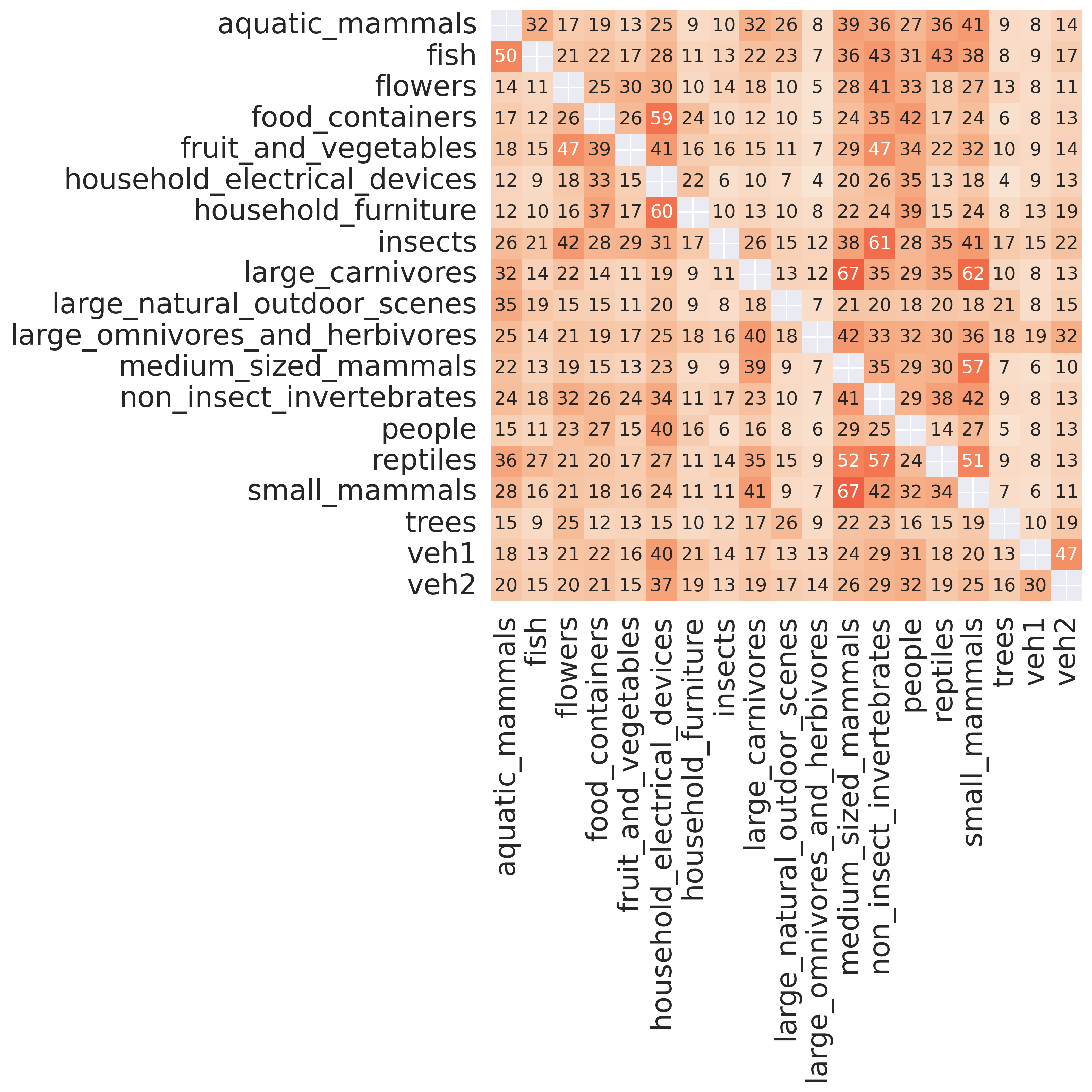}
    \includegraphics[width=0.39\linewidth]{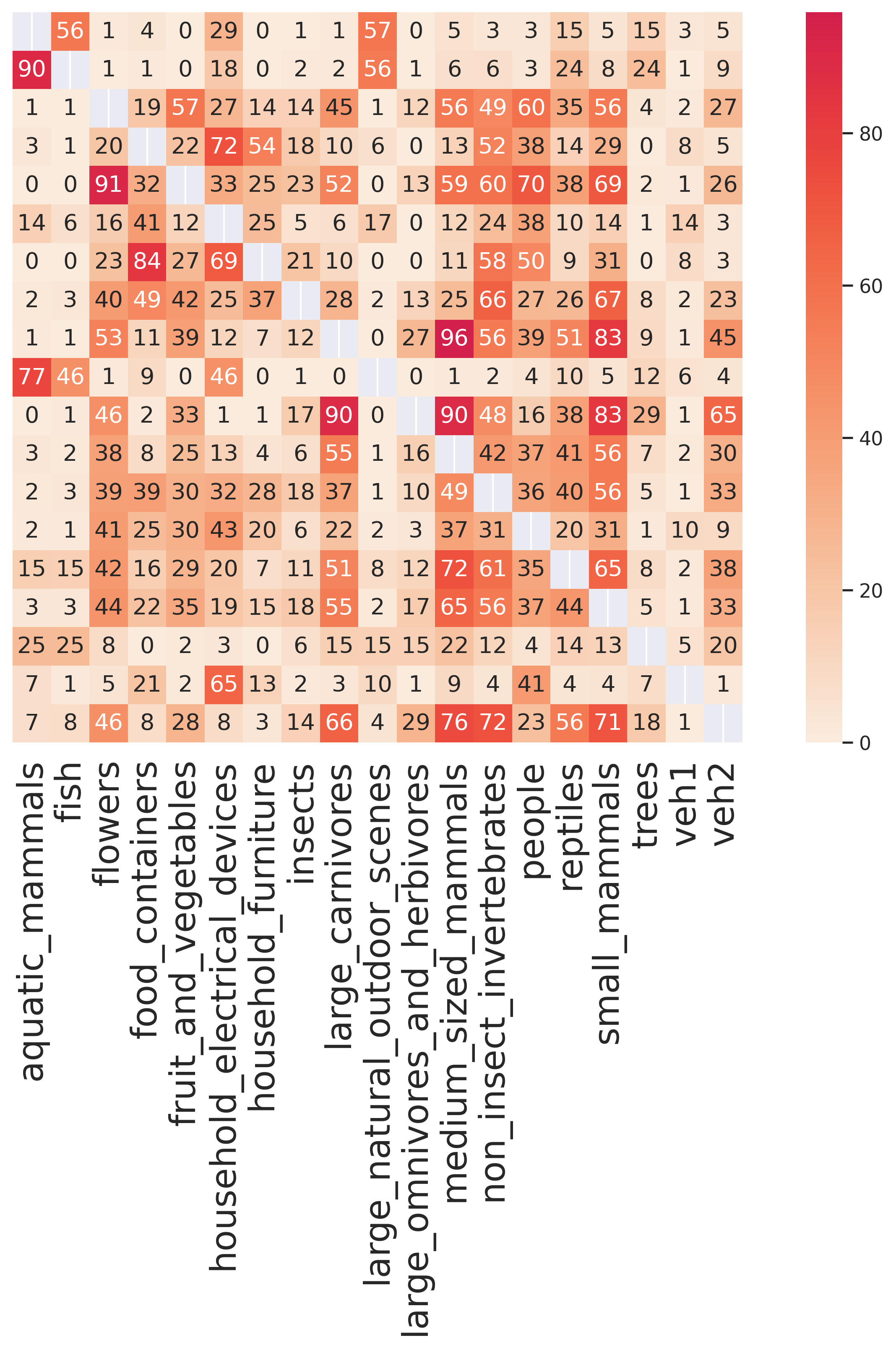}
    \caption{Intersection analysis between selected neurons ranked in the top 
    $12.5\pm6.0\%$
    for trained ViT model (left) and random model (right) in relative importance to Cifar20 classes. Each square shows the percentage overlap between neurons identified as being important for each class pair.}
    \label{fig:cifar20-dot-products}
\end{figure}

In Figure \ref{fig:cifar20-dot-products}, we see that the selected pre-trained neurons have relatively distinct neural activations compared to the randomly initialized model.
We observe that, although there are some differences, the overall pattern of overlaps seems quite similar between the models, albeit with different magnitudes. 

We see three main trends in general going from a random to pre-trained model:
1) More neurons have a closer-to-median overlap.
2) Extreme overlaps were diminished, roughly in proportion to how intuitively dissimilar the classes seem to be.
3) Intuitively similar classes that had little overlap (such as 'vehicles 1' and 'vehicles 2') gain particularly larger overlaps.

\begin{figure}[h]
    \includegraphics[width=0.5\linewidth]{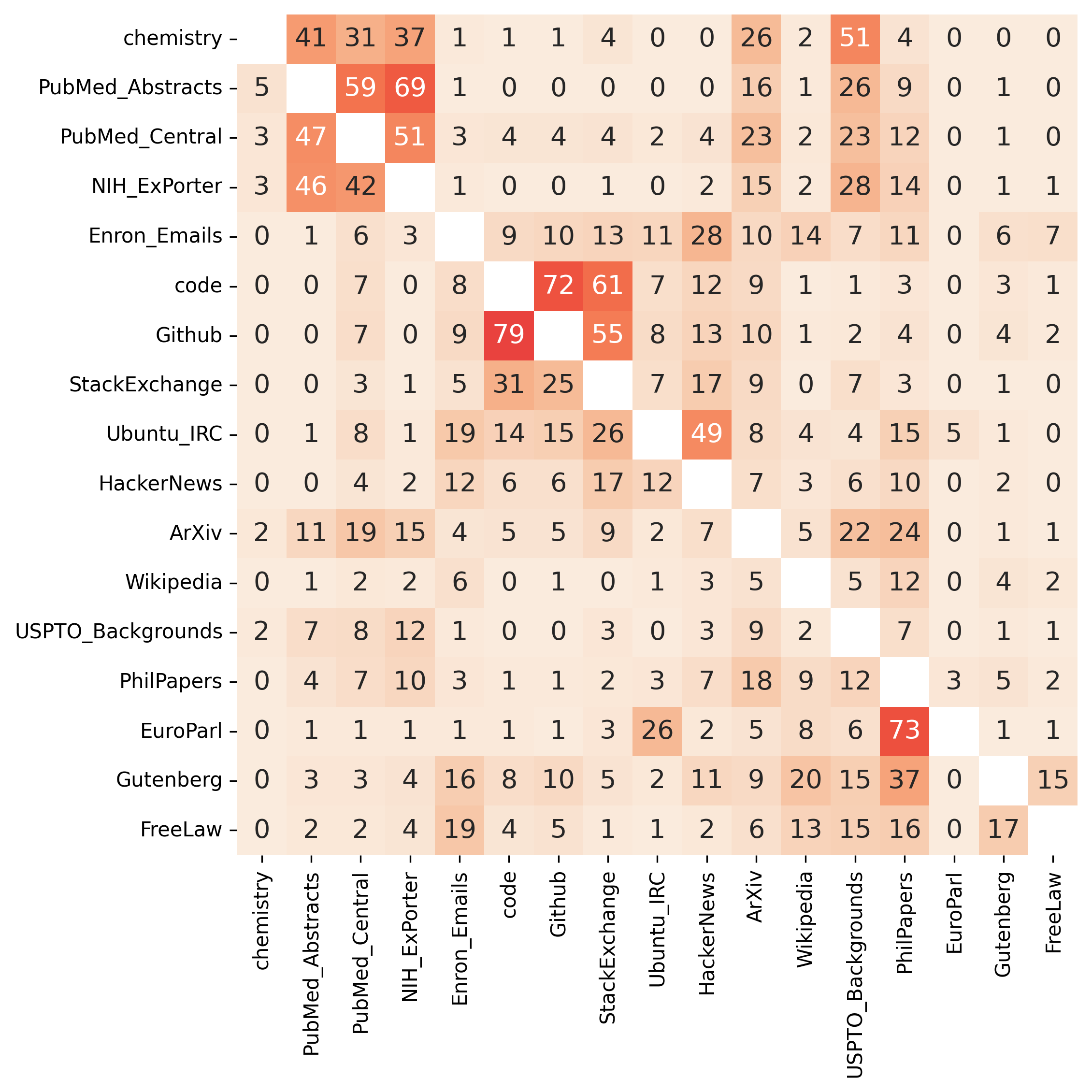}
    \includegraphics[width=0.47\linewidth]{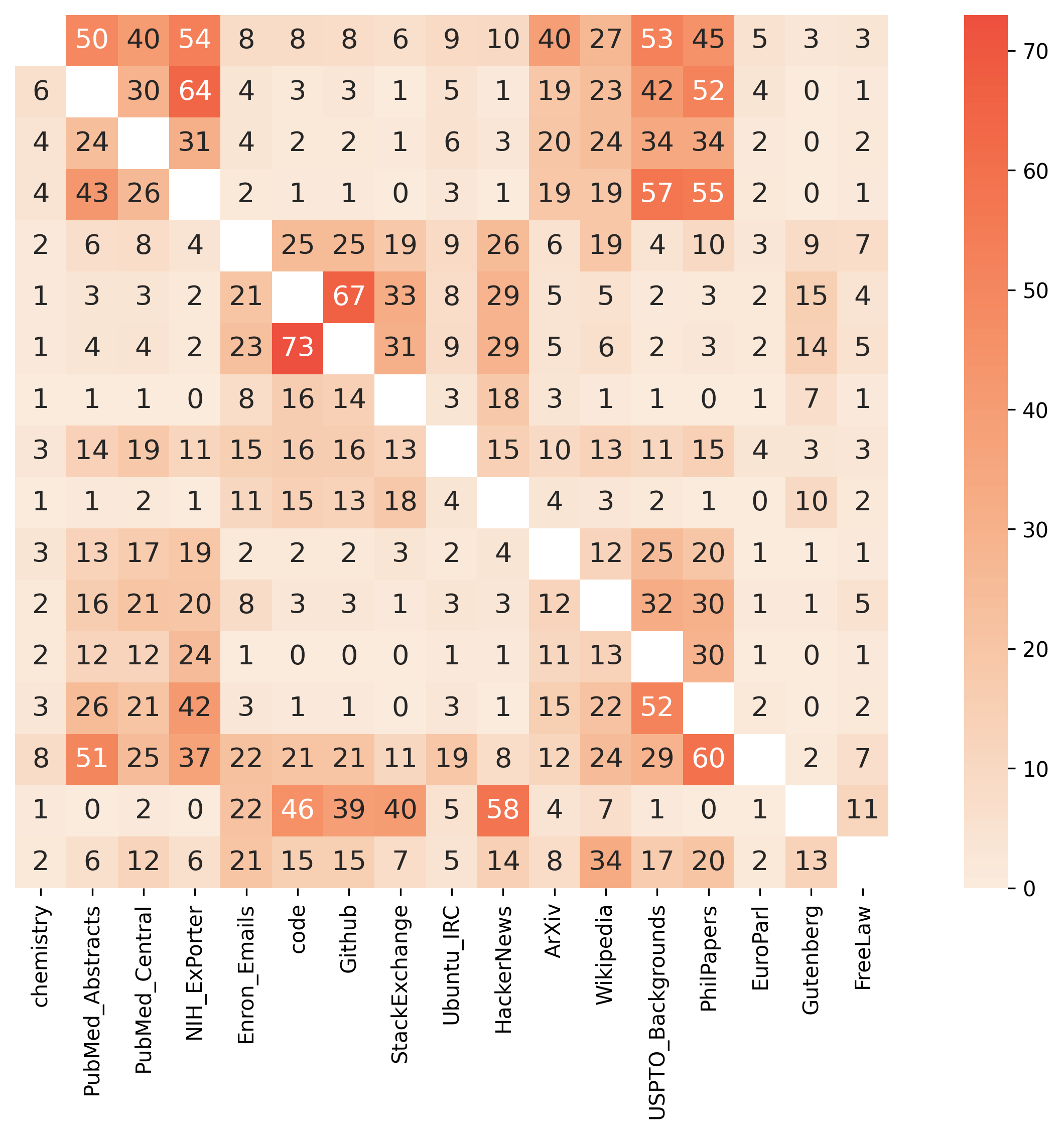}
    \caption{Intersection analysis between selected neurons ranked top 1\% for Mistral 7b model that is trained (left) and randomly initialized (right) in relative importance to different text classes. Each square shows percentage overlap between neurons identified as being important for each class pair.}
    \label{fig:llama-dot-products}
\end{figure}

The results are somewhat analogous for pre-trained and randomly initialized Mistral models seen in Figure \ref{fig:llama-dot-products}. 

We see again three similar trends: 1) More neurons have a closer-to-median overlap, though this time tending to be lower. 2) Most extreme overlaps were diminished, roughly in proportion to how intuitively dissimilar the subsets seem to be. 3) Intuitively similar classes (such as PubMed Abstracts, PubMed Central and NIH ExPorter, or Ubuntu IRC and HackerNews) retain their overlaps or gain larger overlaps.

\begin{figure}[h]
    \centering
    \includegraphics[width=0.49\linewidth]{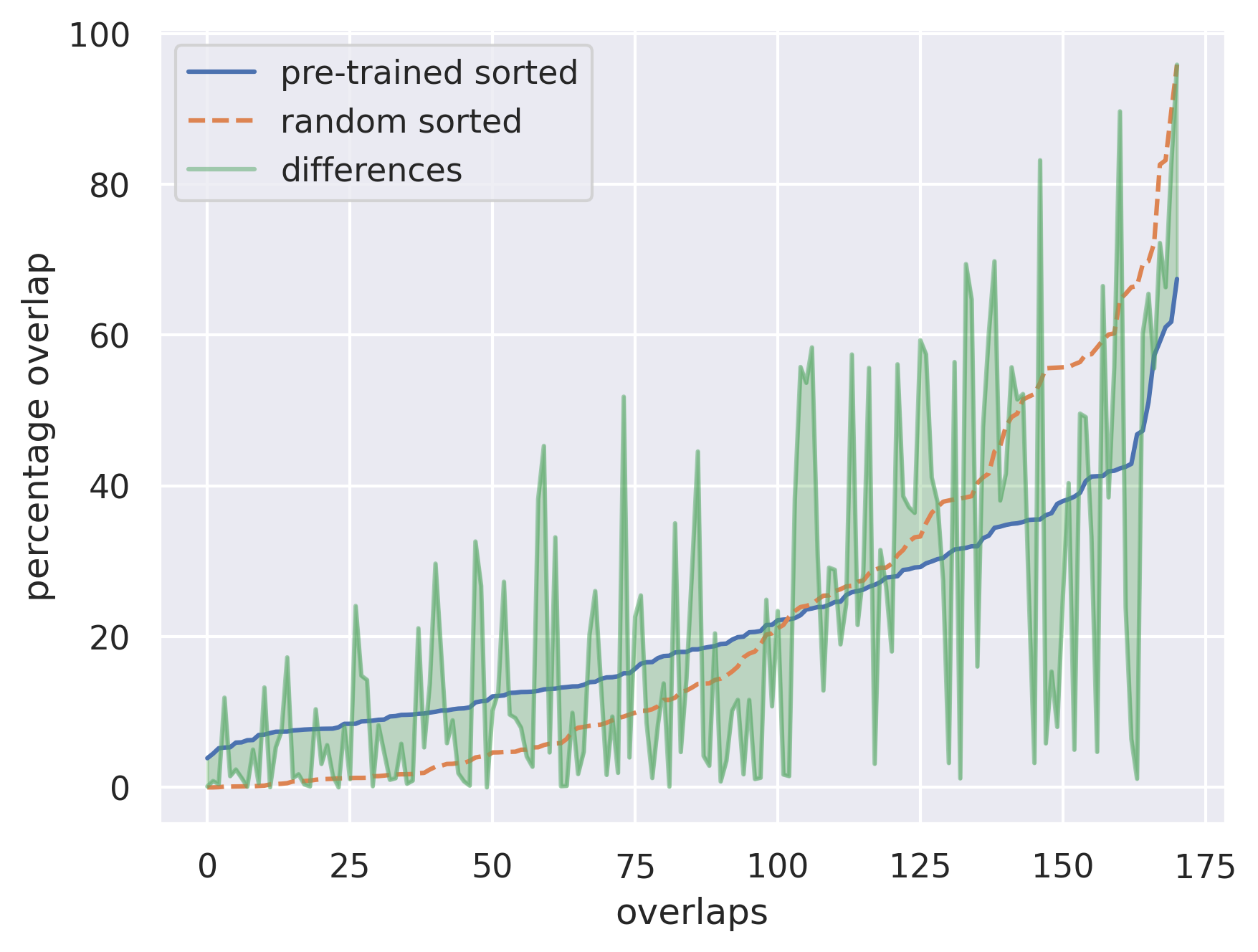}
    \includegraphics[width=0.49\linewidth]{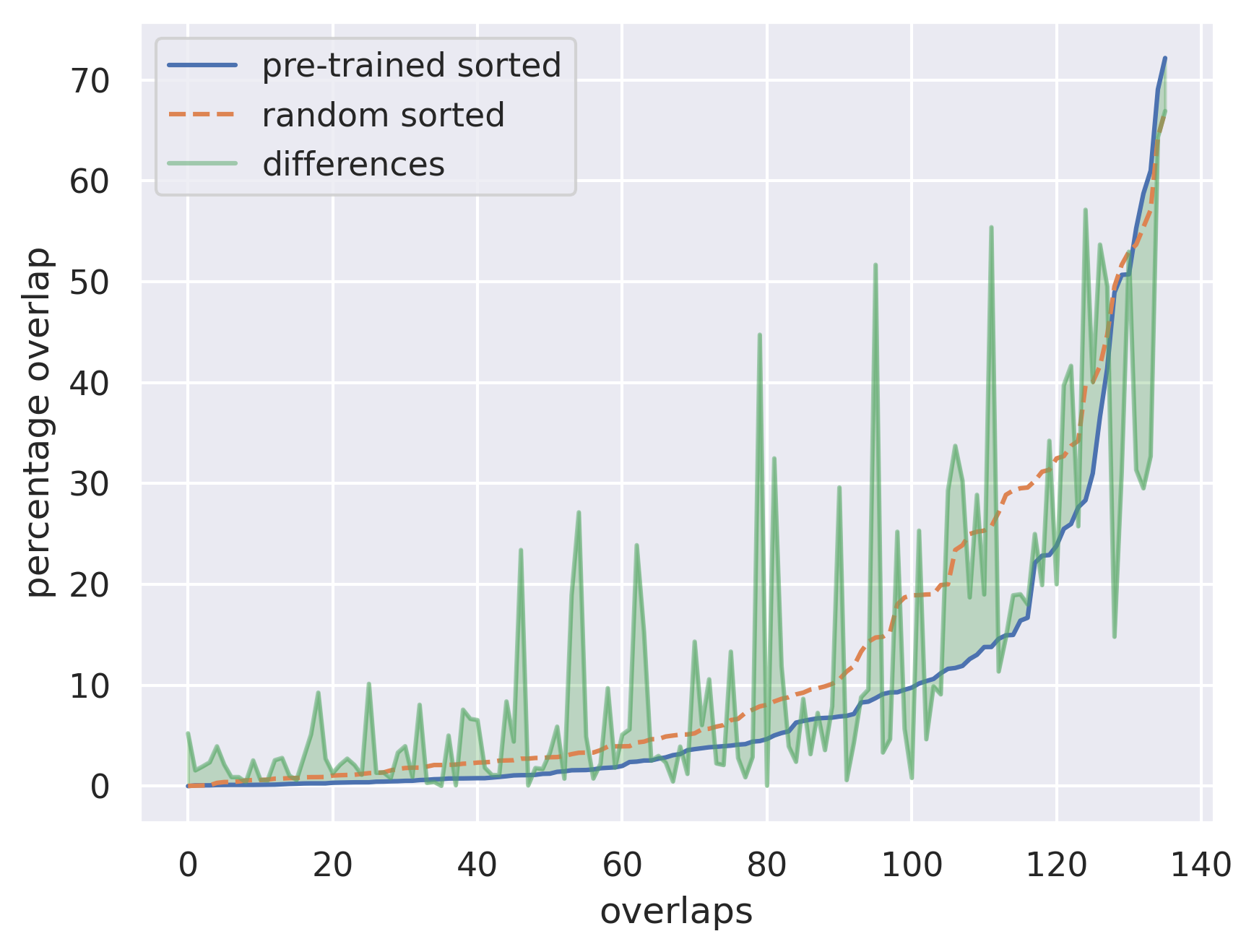}
    \caption{Plot of the overall overlaps between classes of ViT (left) and subtexts of Mistral (right) for both random and pre-trained selected neurons. We also show the difference between the random and pre-trained overlaps at each point.}
    \label{fig:diffs}
\end{figure}

In Figure \ref{fig:diffs}, we see the overlaps between Cifar20 classes and text subsets sorted for both selected pre-trained and selected random neurons. We also see the difference in overlap between pre-trained and random models.
We observe that there is a definite correlation between the overlaps of pre-trained and random neuron selections, but also that there are large shifts. We can more clearly see that the overlaps in ViT are more tightly bound, while in Mistral, the overlaps become overall lower.

Overall, we find the results show a surprising degree to which many of the selected neurons overlap between the randomly initialized and pre-trained models, and see this as evidence for the weak Lottery Ticket Hypothesis \citep{lottery-ticket-hypothesis}.

\subsection{Intersection between Selected Neurons and Clusters}

Inspired by the Lorentz Curve and Gini Coefficient \citep{lorenz-curve-inequality}, for each layer, we measure the degree to which the selected neurons are preferentially selected into few clusters. We do this by counting how many selected neurons are in each cluster, sorting the clusters in order from highest to lowest member count, then plotting a cumulative curve showing distribution between clusters.

\begin{figure}[h]
    \centering
    \begin{subfigure}{.43\textwidth}
        \centering
        \includegraphics[width=\linewidth]{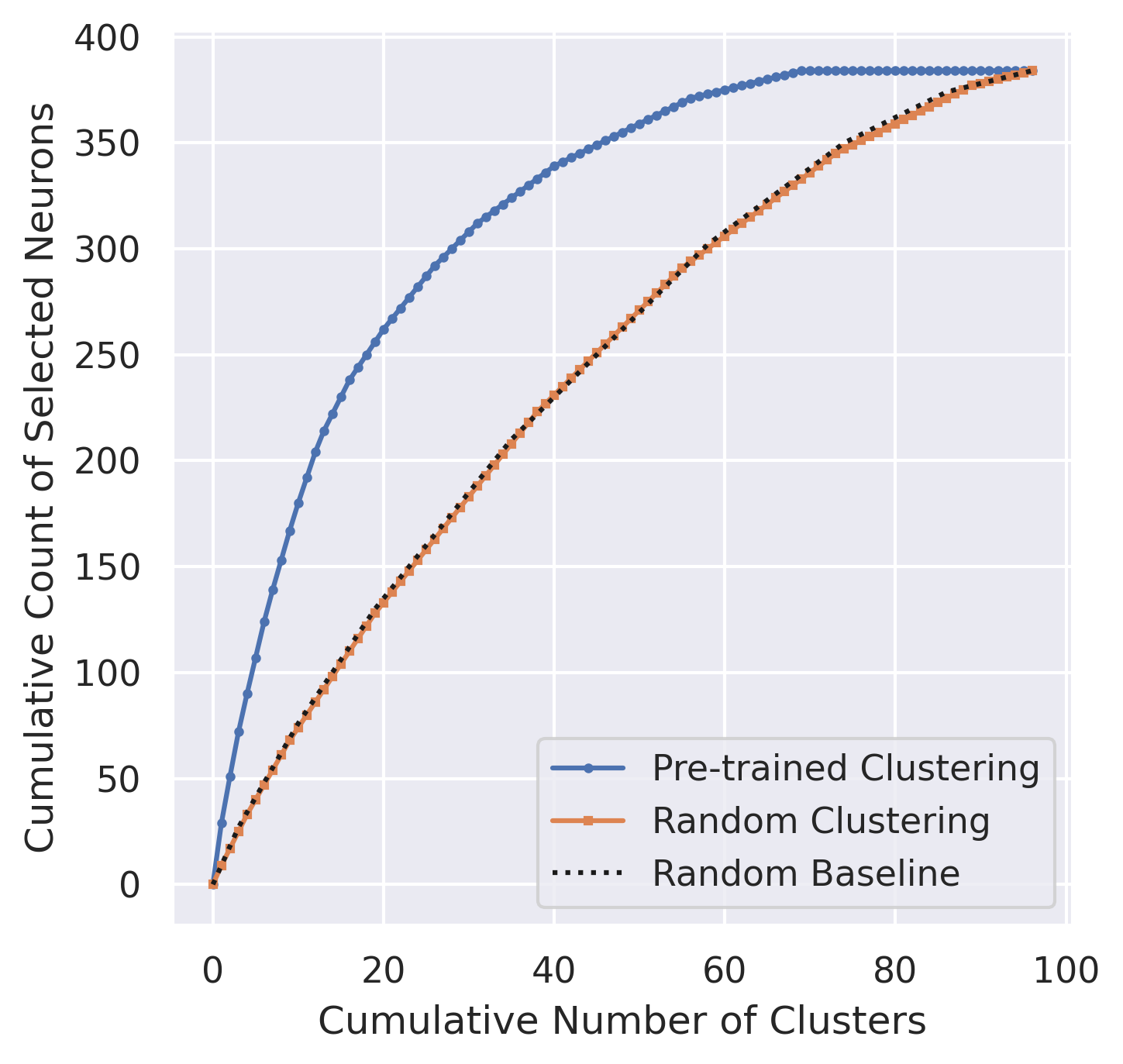}
        \caption{ViT Lorentz Curve}
        \label{fig:vit-lorentz-curve-example}
    \end{subfigure}
    \begin{subfigure}{0.292\textwidth}
        \centering
        \includegraphics[width=\linewidth]{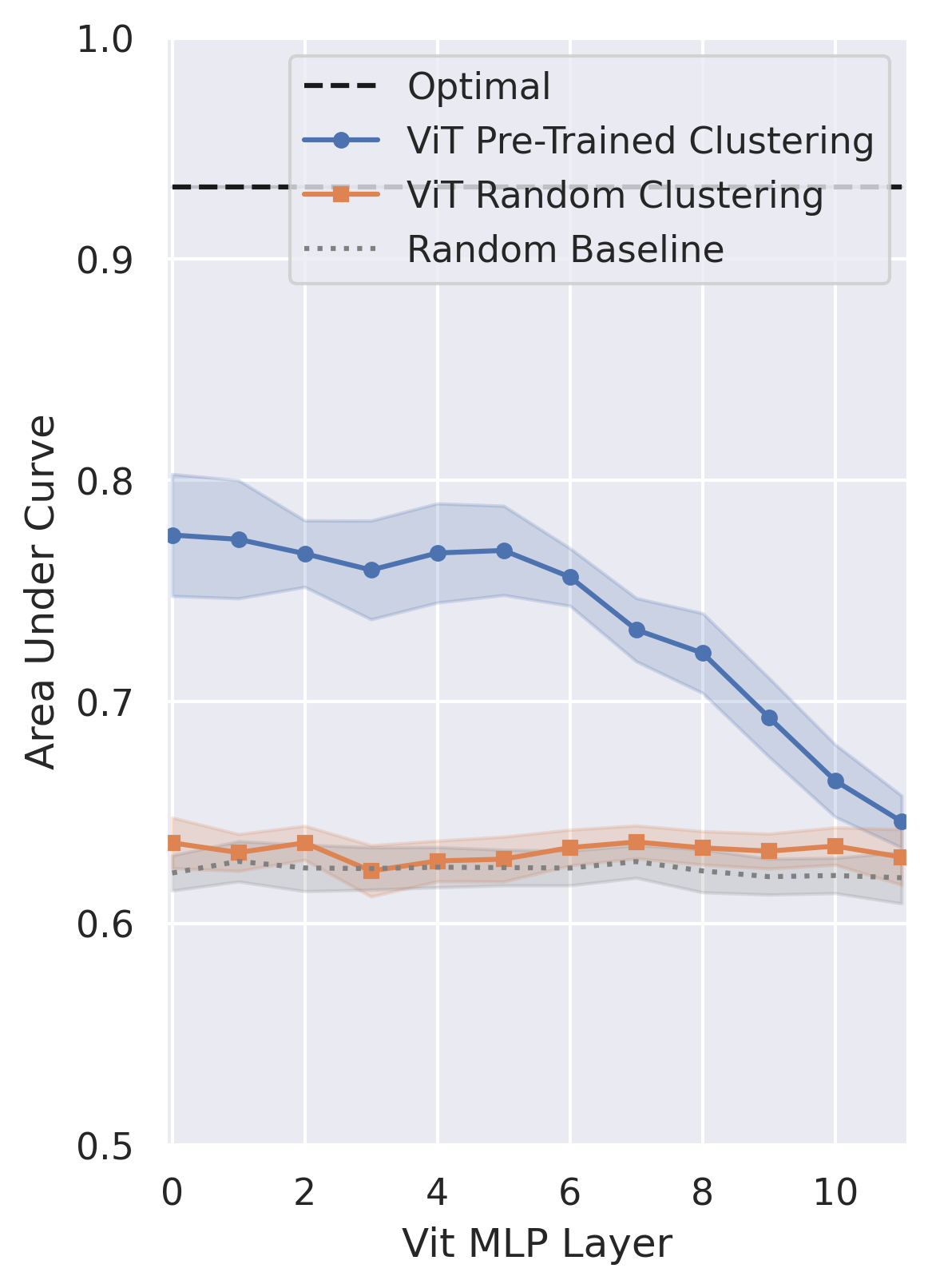}
        \caption{ViT}
        \label{fig:auc-vit}
    \end{subfigure}
    \begin{subfigure}{.252\textwidth}
        \centering
        \includegraphics[width=\linewidth]{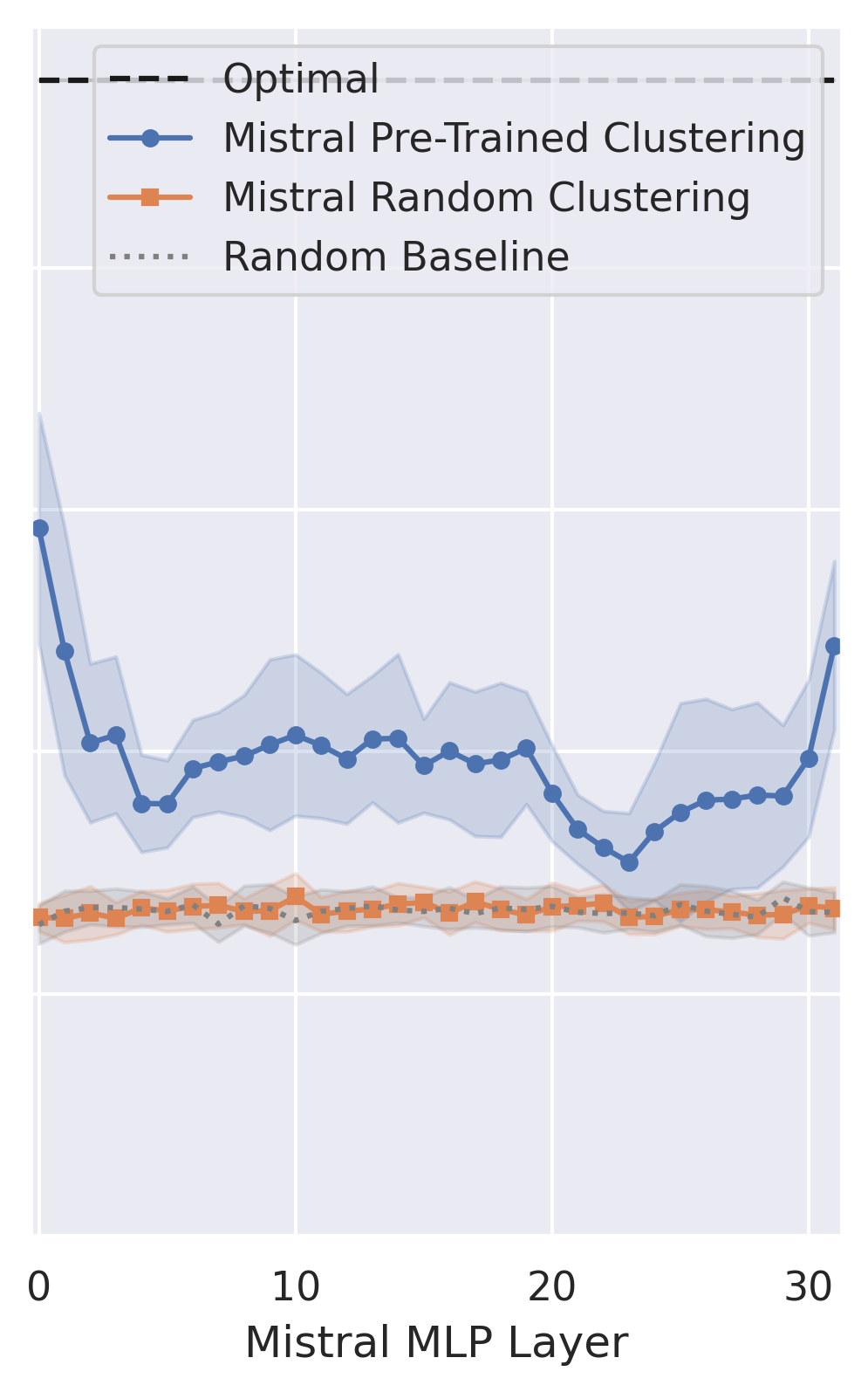}
        \caption{Mistral}
        \label{fig:auc-mistral}
    \end{subfigure}
    \caption{(a) An example of a Lorentz curve in layer 0 of ViT, sorting the distribution of selected top 12.5\% of Cifar20 "people" neurons between MoE clusters in trained and randomly initialized ViT models. The Area Under this Curve (AUC) is then found for all classes of Cifar20 in each layer of ViT (b). The same is done for each data subset of Mistral (c) with the top $3.4\%$ of neurons.}
    \label{fig:enter-label}
\end{figure}

That is, we create $k$ bins, and take the indices $i$ of the selected neurons, and count how many are assigned to each bin, $B[j]$. 
We then sort the bins in order of highest to lowest, giving sorted bins 
$\tilde{B} [j]$,
and plot a cumulative distribution of counts $c_n$, where
$c_n = c_{n-1} + B[n]$. 
We then find the normalized area under the curve, such that if all the neurons were in one bin, the area would be 1.0, and if the neurons were perfectly evenly distributed, it would be 0.5. For an example of such a curve, see Figure \ref{fig:vit-lorentz-curve-example}.

As the MoEfication clustering of neurons is limited to single layers, we add an extra constraint such that the selection of neurons is limited to have an equal proportion per layer, as having all neurons in one layer automatically distributed neurons equally between these clusters.

We see in Figure \ref{fig:auc-vit} that the ViT selected pre-trained neurons correspond relatively strongly with the MoEfication clusters compared to selected random neurons, with stronger correspondence in earlier layers.
In Figure \ref{fig:auc-mistral} we also see that pre-trained neurons correspond more with MoEfication clusters, but this time we see a bi-modal correspondence with later and earlier layers corresponding more than middle layers.

This suggests that the MoEfication in layers close to the text or image seem to map relatively cleanly, but that in more central layers, the co-activation similarity between neurons is not as predictive of class.

\section{Related Work}

\textbf{Circuits-based mechanistic interpretability} seeks to uncover the exact causal mechanisms and neuron activations responsible for specific behaviours \citep{causal-scrubbing,circuits-induction-heads,circuits-ioi-interpretability-redundancy,logit-lens}. Although this bottom-up approach offers detailed insights, the vast number of potential circuits poses a significant challenge to comprehensively understanding model behaviours.

\textbf{Activation-direction-based interpretability} methods, such as Activation Addition \citep{act-add} and Representation Engineering \citep{representation-engineering,dlk-discovering-latent-knowledge,act-add,language-models-space-and-time}, attempt to correlate model activations with meanings or concepts \citep{geva-transformers-in-embedding-space,rome-paper}. However, these top-down approaches often overlook finding the sources of these activation patterns.

\textbf{Neuron-based analysis} has also been explored, with studies suggesting that individual neurons may encode human-interpretable concepts in convolutional neural networks \cite{underthehood}  and language models \citep{DBLP:conf/emnlp/GevaSBL21,geva-promoting-concepts,openai-gpt2-neurons}. However, the complexity and sheer number of neurons, coupled with their potential to represent multiple features, complicate this approach \citep{superposition-toy-models,dropout-superposition}. Techniques like sparse auto-encoder dictionary learning have been proposed to address these challenges, albeit at the cost of increasing the neuron count \citep{superposition-sparse-autoencoders,superposition-sparse-autoencoders-anthropic}.

\textbf{Modularity and Clustering}
Recent research has explored the modular nature of neural networks and how this modularity relates to task specialization. Studies on small neural networks have revealed surprising levels of modularity \citep{modularity-clusterability,modularity-surprisingly-modular-mlps}, while work in computer vision has demonstrated specialized "branches" in convolutional networks \citep{olah-branch-specialization,olah-feature-vis}. In language models, concepts like "MoEfication" \citep{moefication} and contextual sparsity \citep{deja-vu-contextual-sparsity} have shown task-specific neuron activation patterns.

Extending these ideas, \citet{underthehood} characterized learned representations by identifying functional neuron populations in CNNs that either respond jointly to specific stimuli or have similar impacts on network performance when ablated. Their work highlights that a neuron's importance cannot be determined solely by its activation magnitude, selectivity, or individual impact on performance, emphasizing the need for a more nuanced understanding of neural network internals.

Our research builds upon these findings by investigating specialization across different text genres and image classes as a basis for clustering neurons. We adapt selective pruning techniques \citep{pochinkov2024selectivepruning} from Machine Unlearning \citep{machine-unlearning-2021} to assess neuron importance for specific data subsets, aiming to provide new insights into the functional organization of complex neural networks.

\section{Discussion}

Our analysis has shown that both the MoEfication clustering \citep{moefication} and the importance selection from Selective Pruning \citep{pochinkov2024selectivepruning} seem to reveal some degree of interpretable modularity in transformer models. 
We believe this is a promising starting point for understanding the high-level behavior of transformer neural networks, and for identifying how different parts of a neural network might be mapped out, analogously to how regions of a human brain are mapped out in neuroscience.

Additionally, the overlap between similar categories in randomly initialized models seems to provide some evidence for a weak version of the Lottery Ticket Hypothesis \citep{lottery-ticket-hypothesis}.

Our findings have several implications:

\begin{enumerate}
    \item The qualitatively intuitive nature with which classes seem to overlap seems to suggest some level of distinctness with how models handle these inputs. 
    \item The relatively small differences in neuron importance between pre-trained and random models indicate that training does indeed lead to meaningful specialization, but that some of this structure may be present even at initialization.
    \item The clear but incomplete correspondence between MoEfication clusters and task-specific neurons suggests that both methods form a useful but incomplete for understanding models.
\end{enumerate}

However, our study also has limitations that should be addressed in future work:

\begin{enumerate}
    \item We focused on MLP neurons and did not analyze attention mechanisms, which play a crucial role in transformer models.
    \item While we use two very different models and find similar results, our analysis was limited to a specific set of tasks and datasets, and may not generalize to all domains or model architectures.
    \item The interpretation of neuron importance and specialization is still somewhat qualitative and could benefit from more rigorous statistical analysis.
\end{enumerate}

Future work should look to improve the precision with which one can identify different parts of a model that are using different knowledge, as well as increasing the degree to which one can have a fine-grained hierarchical understanding of how the components of a model might be mapped. 

Some potential directions include:
1) Extending the analysis to include attention mechanisms and their interaction with MLP layers.
2) Developing more sophisticated clustering techniques that can capture hierarchical relationships between neurons and tasks.
3) Investigating how neuron specialization evolves during the training process, potentially shedding light on the dynamics of learning in these models.
4) Exploring how this understanding of modularity could be leveraged to improve model interpretability, robustness, or transfer learning capabilities.

\begin{ack}
This study was funded by the Long-Term Future Fund, and conducted in affiliation with AI Safety Camp. The authors would like to thank the anonymous reviewers for their valuable feedback.
\end{ack}

\bibliographystyle{plainnat}
\bibliography{references}

\end{document}